# Estimation of Large Motion in Lung CT by Integrating Regularized Keypoint Correspondences into Dense Deformable Registration

Jan Rühaak*, Thomas Polzin, Stefan Heldmann, Ivor J. A. Simpson, Heinz Handels, Jan Modersitzki, and Mattias P. Heinrich

*Abstract*—We present a novel algorithm for the registration of pulmonary CT scans. Our method is designed for large respiratory motion by integrating sparse keypoint correspondences into a dense continuous optimization framework. The detection of keypoint correspondences enables robustness against large deformations by jointly optimizing over a large number of potential discrete displacements, whereas the dense continuous registration achieves subvoxel alignment with smooth transformations. Both steps are driven by the same normalized gradient fields data term. We employ curvature regularization and a volume change control mechanism to prevent foldings of the deformation grid and restrict the determinant of the Jacobian to physiologically meaningful values. Keypoint correspondences are integrated into the dense registration by a quadratic penalty with adaptively determined weight. Using a parallel matrix-free derivative calculation scheme, a runtime of about 5 min was realized on a standard PC. The proposed algorithm ranks first in the EMPIRE10 challenge on pulmonary image registration. Moreover, it achieves an average landmark distance of 0.82 mm on the DIR-Lab COPD database, thereby improving upon the state of the art in accuracy by 15%. Our algorithm is the first to reach the inter-observer variability in landmark annotation on this dataset.

## I. INTRODUCTION

THE registration of inspirative and expirative lung CT images has important medical applications, in particular

*J. Rühaak is with Fraunhofer MEVIS, Institute for Medical Image Computing, 23562 Lübeck, Germany (e-mail: jan.ruehaak@mevis.fraunhofer.de).

T. Polzin is with the Institute of Mathematics and Image Computing, University of Lübeck, 23562 Lübeck, Germany.

S. Heldmann is with Fraunhofer MEVIS, Institute for Medical Image Computing, 23562 Lübeck, Germany.

I. J. A. Simpson is with Anthropics Technology Ltd., London W12 7SB, U.K.

H. Handels and M. P. Heinrich are with the Institute of Medical Informatics, University of Lübeck, 23562 Lübeck, Germany.

J. Modersitzki is with Fraunhofer MEVIS, Institute for Medical Image Computing, 23562 Lübeck, Germany, and also with the Institute of Mathematics and Image Computing, University of Lübeck, 23562 Lübeck, Germany.

in the diagnosis and characterization of chronic obstructive pulmonary disease (COPD) [1]. COPD is characterized by chronic airflow limitation, which is primarily caused by small airway disease and emphysema [2]. Using accurate pulmonary registration, the local lung ventilation can be reliably quantified [3]. Furthermore, lung registration is widely used in radiotherapy for estimation of tumor motion [4] together with localized ventilation measures, which can be included into planning in order to spare as much well-functioning tissue as possible from radiation dose. Local volume change can be calculated based on the Jacobian determinant of nonlinear displacement fields [5].

From our experience, the following two conditions are most important for an accurate motion estimation: 1) correct alignment of all discriminative inner-lung structures (such as vessels and airways); 2) smooth and plausible (in particular invertible) deformations with subvoxel accuracy. While several approaches have been proposed in the past that aim at optimizing both objectives jointly (see [6] for an overview), there exists a remaining challenge to achieve highly accurate results for scan pairs between full inspiration and full expiration and therefore very strong motion [7]. Here we propose to decouple the complex problem into two steps: 1) a very robust computation of sparse correspondence fields for a moderate number of keypoints and 2) an intensity-driven, continuous-optimization based deformable registration integrating both keypoint correspondences and volume change constraints. This work draws inspiration from our two previous conference publications [8], [9]. In contrast to heuristic approaches that use a separate descriptor matching, we integrally combine keypoint correspondences and continuous optimization within one unified objective function. We present improved pre-processing, data-adaptive parametrization and parallel matrix-free continuous optimization together with a substantially extended evaluation on publicly available lung CT images.

### A. Related Work and Motivation

There are several competing strategies for intra-subject alignment of volumetric lung CT scans. Following [6], their performance mainly differs based on their choice of energy terms and optimization strategy.

*1) Similarity Metrics:* One of the main challenges of pulmonary image registration is the reduction in lung density during inspiration due to inflowing air. The associated lung volume expansion leads to a decrease of the Hounsfield units in the parenchymal region, violating the intensity constancy assumption between corresponding points, upon which the classic sum of squared differences (SSD) distance measure is built. Several alternatives are reported in the literature: In [10] and [11], local cross-correlation is used as distance measure for robustness against breathing-related changes. A mass-preserving transformation model based on the computed volume change is used in [12] to adapt the local image intensity, enabling the use of SSD without violating the intensity constancy assumption. A similar approach has been pursued in [13] by using the so-called sum of squared tissue volume differences (SSTVD) distance measure together with a vesselness filter for improved alignment of vascular structures. In our work, normalized gradient fields (NGF) [14] are used to gain invariance against changes in tissue density.

*2) Optimization Strategy:* Since lung registration yields a highly nonlinear and usually non-convex problem, continuous optimization approaches may easily get trapped in local minima. In addition to specific pre-processing protocols (e.g. alignment of lung segmentations and masking of outer-lung tissue, see Section II-A), most approaches attempt to deal with this problem by employing a multi-resolution framework. However, for large intra-patient motion between inspirative and expirative scans residual misalignments are still common. This is evident from results published in [15] when applying three widely used registration toolboxes on the COPD dataset of [7], resulting in relatively high average landmark errors of 1.58 mm [16], 4.68 mm [17] or 2.19 mm [18]. When employing discrete optimization, a large range and dense sampling of potential displacements can be explored simultaneously (as in e.g. [8], [19], [20]), reducing average errors to 1.08 mm [8].

Our strategy is therefore to sample as many displacements as possible to find robust correspondences with the help of discrete optimization, while reducing the computational complexity by considering only a sparse subset of keypoints (as control points) and employ similar approximations for the Markov random field based regularization as in [20]. Invertibility and volume change constraints of the estimated transformation, which are difficult to model in discrete optimization and computationally expensive for a large number of displacements, are only enforced in the subsequent dense registration to ease the keypoint matching.

Deformable registration with an initial keypoint correspondence search has been considered in previous work [21]–[24]. However, the matching of each keypoint has been performed independently, which thus generally resulted in a considerable number of outliers. For inter-subject brain registration, a graph-based matching of two sets of landmarks was proposed in [25] and [26] that optimizes all correspondences simultaneously. In most approaches, the employed feature descriptors are not applicable as a similarity metric in the subsequent dense continuous optimization, yielding an inconsistent registration model. Our proposed method optimizes jointly over all potential combinations of keypoint correspondences using a graph-based optimization technique. In contrast to [22], [25], [26], we detect keypoints in one scan and use a dense search window in the other. Adding spatial regularization yields more robust correspondences as shown in [8]. Part-based models [27] have also been successfully applied to anatomical landmark localization [28]. Here, they are used for the first time as energy term within an intensity-driven deformable registration.

## B. Paper Outline

In Section II our proposed lung registration method DIS-CO is described. The images are first preprocessed using lung masks to account for large scalings and to exclude non-lung voxels. Second, a sparse correspondence field is found using a part-based model and the normalized gradient fields distance measure. These keypoint matches are then integrated as least-squares penalty into a continuous optimization, which is regularized by a curvature term and a volume change control mechanism. In Section III the presented method is extensively evaluated on publicly available lung CT data and compared to state-of-the-art algorithms in terms of landmark and fissure distances as well as the Jacobian determinant. The proposed algorithm ranks first in the EMPIRE10 challenge [6] and achieves an average landmark distance of 0.82 mm on the DIR-Lab COPD database [7], thereby improving upon the state of the art in accuracy [29] by 15 %. Our algorithm is the first to reach the inter-observer variability in landmark annotation on this dataset as highlighted in Section IV.

## II. METHODS

Let $\mathcal{F} : \mathbb{R}^3 \rightarrow \mathbb{R}$ denote the fixed image and $\mathcal{M} : \mathbb{R}^3 \rightarrow \mathbb{R}$ the moving image with compact support in domains $\Omega_{\mathcal{F}}$ and $\Omega_{\mathcal{M}}$, respectively. We aim to estimate a nonlinear transformation $y : \mathbb{R}^3 \rightarrow \mathbb{R}^3$ that best aligns $\mathcal{F}$ (the inspiration CT scan) and $\mathcal{M}$ (an expiration scan of the same patient).

## A. Preprocessing

Prior to our intensity-based registration, the CT scans are preprocessed using lung segmentations. In our experiments, we employ the lung segmentation algorithm from [30]. Following [31], the lung masks are first aligned by their centers of gravity, followed by an affine-linear registration using the SSD distance measure on these binary masks. In order to fully exploit the segmentation information and provide a better alignment of the deformable lung surface at different breathing states, we extend the model of [31] as follows. An additional deformable registration [32] of the lung segmentations is performed using again the SSD distance of lung masks together with curvature regularization [33] and a penalty on local volume change [34]. This mimics our intensity-driven registration framework as described in detail in Section II-D1 and yields an initial transformation $\hat{y}$ together with a pre-aligned moving image $\hat{\mathcal{M}}$.

We further use the lung segmentation to mask both CT images, thus restricting the deformation to the lung area of interest as illustrated in Fig. 1. This removes the necessity of recovering sliding motion between lung and rib cage during breathing, more precisely between *pleura visceralis* and *pleura parietalis*, which otherwise requires additional attention in the deformation modeling [35]–[38]. Note that in many clinical applications, including lung cancer screening and quantitative lung ventilation analysis, a lung segmentation mask is usually available for both scans. The masked lung scans are then used as input to all subsequent steps, cf. also Fig. 1.

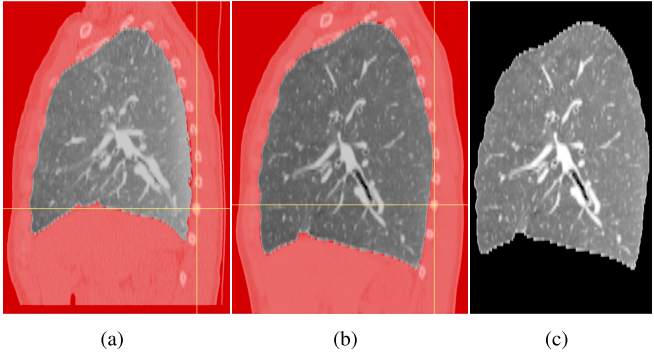

Fig. 1. Sagittal views of dataset COPD6: **(a)** exhale and **(b)** inhale scan with a red overlay of the inverse lung mask. Only the intensities within the lungs are retained, as shown for the inhale scan in **(c)**. Note the volume change that is indicated by the yellow crosshair located at the same rib in both scans. The increased Hounsfield units in the parenchyma and the deteriorated quality due to reduced acquisition dose [7] of the exhale scan are visible.

### B. Normalized Gradient Fields

Focusing on image edges rather than on absolute intensities is particularly beneficial for the registration of inspirative and expirative lung CT scans, since the inflowing air may easily lead to differences of several hundred Hounsfield units between corresponding parenchymal structures [7]. Major edges are given by the bronchial and vessel tree, the lung boundary and the fissures. For the alignment of such structures, we employ the following variant of the normalized gradient fields (NGF) distance measure [9], [14]

$$\mathcal{D}(\mathcal{F}, \mathcal{M}(y)) := \int_{\Omega_{\mathcal{F}}} 1 - \frac{\langle \nabla \mathcal{M}(y(x)), \nabla \mathcal{F}(x) \rangle_{\eta}^2}{\|\nabla \mathcal{M}(y(x))\|_{\eta}^2 \|\nabla \mathcal{F}(x)\|_{\eta}^2} \, dx \quad (1)$$

with $\langle f, g \rangle_{\eta} := \eta^2 + \sum_{j=1}^3 f_j g_j$ and $\|f\|_{\eta}^2 := \langle f, f \rangle_{\eta}$. It utilizes image edges independently of their strength, but allows to suppress small noise-related edges by an edge parameter $\eta > 0$. A key element of this work is that the same cost term (NGF) is used throughout both discrete keypoint matching and dense continuous optimization. Note that we do not claim NGF to be superior to other contrast-invariant metrics, its choice is motivated by its comparably efficient computation, robustness, and suitability for numerical optimization.

### C. Regularized Keypoint Correspondences

Keypoints are widely used in image recognition [39] and multi-view scene reconstruction [40]. In order to cope with large motion, searching for sparse keypoint matches has been proposed in previous work [21]–[23]. However, these approaches have in common that an unconstrained optimum is found for each keypoint independently, potentially leading to outliers. Our proposed approach for inferring regularized correspondence fields, which has been initially outlined in [8], consists of three main parts: sparse keypoint extraction, similarity evaluation over search space, and a part-based model for inference of a regularized transformation. This is followed by a refinement stage, which takes into account the symmetry of displacement marginals (see Section II-C4). To avoid repetitive interpolation, both scans are resampled to an isotropic resolution of 1 mm$^3$ for keypoint matching.

*1) Sparse Keypoint Extraction:* As suggested by previous approaches for interest point localization in lung CT scans [21], [23], a Förstner operator [41] is applied to find a sparse set of distinctive keypoints $K \subset \mathbb{R}^3$ in the fixed image. The spatial gradients of the fixed scan $\nabla \mathcal{F}$ are smoothed with a Gaussian kernel $G_{\sigma}$, which yields a distinctiveness volume $F_D = 1/\operatorname{trace} \left( (G_{\sigma} * (\nabla \mathcal{F} \nabla \mathcal{F}^T))^{-1} \right)$. To avoid strong spatial clustering of detected keypoints, a maximum filter over a 3D cubic region $R$ is applied to the Förstner response: $F_D^* = \max_{x \in R} F_D(x)$. Locations that are not a local maximum, i.e. where $F_D \neq F_D^*$, are removed from $K$. Since we are interested in aligning inner lung structures, the keypoint extraction is restricted to the lung volume. By empirically choosing $\sigma = 1.4$ mm and a side-length of 6 voxels for $R$, we obtain a set of $|K| \approx 3500$ well-dispersed keypoints.

*2) Similarity Evaluation Over Search Space:* Matching keypoints has been proposed based on SIFT-like feature vectors [42] for natural images [22], [43] and medical scans [24] or by learning appropriate Gabor features in [44]. Here, we refrain from using such computationally expensive descriptors and simply employ normalized gradient fields (see Section II-B) motivated by the fact that they are robust against local intensity variations and strong acquisition noise, which is common in ultra-low-dose CT. However, it is expectable that other low-complexity descriptors (e.g. Census as used in [19] or self-similarity descriptors (SSC) [8], [45]) yield a comparable performance. We found in previous work [46] that, due to the difficulty of detecting the same interest point twice across medical scans, classical keypoint-to-keypoint matching [22] may yield insufficient correspondence field quality. We therefore extract keypoints only in the fixed image and search for potential matches over a large range of finely quantized displacements $d \in \mathcal{L} = \{0, \pm 2, \pm 4, \ldots, \pm 32\}^3$ mm with $|\mathcal{L}| = 35937$. Our distance matching cost $\mathcal{D}_{\mathrm{KP}}(k, d)$ (dependence on images $\mathcal{F}$ and $\hat{\mathcal{M}}$ omitted for brevity) is defined as sum of pointwise normalized gradient costs (see (1)) for a small patch $P_k$ of voxels $p$ around a keypoint $k \in K$ and a certain new location $k + d$ within the pre-aligned moving image $\hat{\mathcal{M}}$:

$$\mathcal{D}_{\mathrm{KP}}(k, d) := \frac{1}{|P_k|} \sum_{p \in P_k} 1 - \frac{\langle \nabla \hat{\mathcal{M}}(p + d), \nabla \mathcal{F}(p) \rangle_{\eta}^2}{\|\nabla \hat{\mathcal{M}}(p + d)\|_{\eta}^2 \|\nabla \mathcal{F}(p)\|_{\eta}^2}. \quad (2)$$

Thus for each of the $|\mathcal{L}|$ potential displacements the matching likelihood is computed by summing $|P_k|$ NGF costs. We used a patch-size of $7^3$ voxels with a stride of 2, yielding $|P_k| = 64$. A traditional block-matching approach [23], [47], [48] would then select an optimal displacement for each keypoint $k$ independently: $d^* = \arg\min_{d \in \mathcal{L}} \mathcal{D}_{\mathrm{KP}}(k, d)$. However, this procedure is prone to outliers, thus making (often ad-hoc) post-processing schemes necessary to eliminate erroneous matches, as qualitatively demonstrated in Fig. 2(a). In contrast, we will leverage the advantages of discrete optimization that has recently been popularized in medical image registration [17], [20] in order to find the optimal combination of displacement vectors jointly for all sparse keypoints.

*3) Part-Based Model for Inference of Regularization:* In contrast to traditional block-matching approaches that assume independence of neighboring motion vectors, our goal is to find a sparse regularized transformation $y$ that combines a prior assumption of globally smooth lung deformations with the local displacement likelihoods $\mathcal{D}_{\mathrm{KP}}(k, d)$, estimated in the previous section. Given their success in deformable object and pose recognition, we propose to use part-based models [27] to find the optimal joint configuration of displacements of parts (here lung keypoints), which are connected within a graph. Including this regularization into the matching of sparse keypoints removes the need for complex image descriptors to obtain robust and accurate alignment of highly deformable lung structures and enables the use of the same similarity term (NGF) as in the following dense intensity-driven step.

Ideally, the regularization term would also be identical to the curvature term (5) used within our continuous optimization part, yet realizing this within a discrete optimization model requires the computationally demanding use of ternary (higher-order) cliques instead of pairwise potentials [49]. We therefore opt to use an efficient approximation based on a diffusion-like regularizer $\mathcal{R}_{\mathrm{KP}}(d_k, d_q)$:

$$\mathcal{R}_{\mathrm{KP}}(d_k, d_q) := \frac{\alpha_{\mathrm{KP}} ||d_k - d_q||_2^2}{||x_k - x_q||_2 + |\mathcal{F}(k) - \mathcal{F}(q)|/\sigma_I}. \quad (3)$$

Here, $\alpha_{\mathrm{KP}}$ is a user-determined weighting term; the squared differences of displacements of two neighboring keypoints $k$ and $q$, which are connected by an edge $e_{kq}$, are penalized (normalized by their spatial and intensity distance, with $\sigma_I = 150$). Minimizing $\sum_{k \in K} \mathcal{D}_{\mathrm{KP}}(k, d_q) + \sum_{e_{kq} \in E} \mathcal{R}_{\mathrm{KP}}(d_k, d_q)$ for an arbitrary graph is still NP-hard, we thus restrict the graph to a minimum-spanning-tree (where the denominator of (3) describes the edge cost between keypoints) as commonly done in part-based models [27], [28]. Belief propagation is used to obtain the exact marginal energies for every keypoint and displacement label after two passes of messages. Starting from the leaf nodes, messages $m$ are passed along the edges $e_{kq} \in E$ of the tree (from current node $k$ to parent node $q$) and updated with the following computation of one element of $m_{kq}$ [27]:

$$m_{kq}(d_q) = \min_{d_k} \mathcal{D}_{\mathrm{KP}}(k, d_q) + \mathcal{R}_{\mathrm{KP}}(d_k, d_q) + \sum_c m_{ck}(d_k),$$
$$(4)$$

where $c$ are the children of $k$. In theory, every potential combination of two displacements $d_k$ (from the current node $k$) and $d_q$ (its parent $q$) would have to be calculated. However, for the regularizer chosen in (3), the computation of each message vector $m_{kq}$ is particularly efficient (linear in $|\mathcal{L}|$) when using distance transforms [27]. Once the root node has been reached, messages have to be passed in the opposite directions to obtain

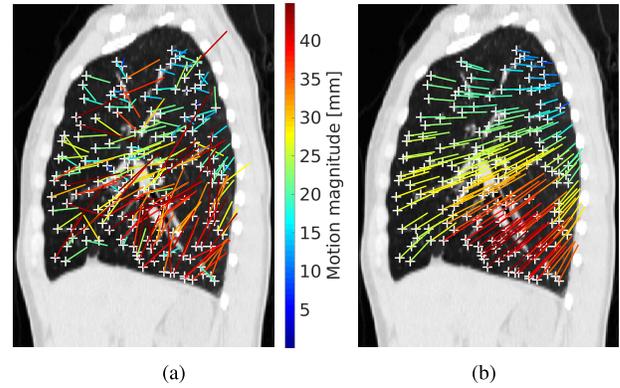

(a)             (b)

Fig. 2. Sagittal views of dataset COPD6 [7]: Scan with keypoint correspondences generated either (a) without regularization or (b) with regularization. The lines show the direction of the movements of keypoints during respiration and their color indicates the magnitude of the motion.

a vector of marginal energies for each node (or keypoint). More details on this outward pass can be found in [50].

*4) Refinement by Symmetry of Marginal Distribution:* Despite the part-based model regularization, the estimated correspondences could still be erroneous due to the repetitive appearance of different lung structures within the same search window, as well as the slightly limited regularization given by the tree model. To overcome these difficulties, we perform an additional estimation of the marginal distribution in the opposite direction. Similar to a mutual consistency check often employed in stereo-depth processing [51], we first translate the keypoint locations according to their regularized matches into the moving image and use them as control points for a second correspondence estimation (following the same procedure as in Sections II-C2 and II-C3) from moving to fixed image. We then simply average the respective marginal energies, see [8] for more details. Ideally, the energies of both directions of the displacement space are symmetric and the averaging re-enforces them. If the displacement from fixed to moving image pointed to a close neighbor of the correct displacement and the motion in a small neighborhood is smooth, the backward search may improve the match.

Afterwards, a new dense transformation could be directly obtained by choosing the minimizer $d_k^*$ of the marginal energies for each keypoint and some interpolation strategy (in our previous work thin-plate spline transforms were used [8], [52]). The relaxed regularization term (3) and the complex motion may, however, still cause such a transformation to contain implausible foldings (negative Jacobian determinants) and is therefore problematic as initialization for a continuous optimization strategy that might not recover these errors. We therefore opt for a sparse correspondence field $y_{\mathrm{KP}}$, which is only defined at keypoint locations, and use it for a keypoint penalty $\mathcal{K}$ as detailed later in Section II-D2. The effect of the regularization on the keypoint matching and the resulting sparse correspondence fields are shown in Fig. 2.

### D. Dense Intensity-Driven Registration Model

The employed dense intensity-driven registration algorithm is based on the lung registration method from [9]. Here,

a standard variational approach [32], [53], [54] is extended by two terms, one that improves lung boundary alignment [34] and one controlling local volume change, thereby restricting the transformation to meaningful volume change and guaranteeing invertibility, cf. also [55]. We augment the method by three elements: an extended pre-registration using nonlinear transformations (cf. Section II-A), integration of keypoint correspondences as an additional term in the objective function, and an adaptive parameter estimation strategy. Moreover, parallel matrix-free schemes for derivative calculations [56] were employed in all registrations, thereby reducing runtime and memory consumption.

The classic variational registration approach, which forms the basis of our algorithm, models the transformation $y$ as a minimizer of a joint objective functional $\mathcal{D}(\mathcal{F}, \mathcal{M}(y)) + \alpha \mathcal{R}(y)$ with regularization parameter $\alpha > 0$. As stated before, we employ the NGF distance measure (1) for $\mathcal{D}$. Our regularization strategy is founded on the observation that non-continuous sliding motion along the rib cage does not have to be recovered when using lung segmentations. Hence, the remaining motion inside the lungs is expected to be very smooth, which motivates the usage of the curvature regularizer [33]

$$\mathcal{R}(y) := \frac{1}{2} \int_{\Omega_{\mathcal{F}}} \sum_{j=1}^{3} \| \Delta(y_j - \hat{y}_j) \|^2 \; \mathrm{d}x, \qquad (5)$$

with a given transformation $\hat{y}$. The curvature regularizer penalizes second order derivatives of the deviation of $y$ from $\hat{y}$, thus yielding very smooth deformations. We set $\hat{y}$ to the result of the pre-registration of the lung masks, cf. Section II-A.

*1) Model Extension:* Following [9], the basic registration model is extended by two terms. As a first fundamental requirement, the computed deformation shall map tissue within the lung in scan $\mathcal{F}$ to tissue within the lung in scan $\mathcal{M}$, and analogously the area outside the lung in scan $\mathcal{F}$ to the area outside the lung in scan $\mathcal{M}$. To this end, let $b_{\mathcal{F}} : \Omega_{\mathcal{F}} \to \{0, 1\}$ and $b_{\mathcal{M}} : \Omega_{\mathcal{M}} \to \{0, 1\}$ denote binary functions for $\mathcal{F}$ and $\mathcal{M}$ that are equal to one inside the lungs and zero otherwise. We define the penalty term $\mathcal{B}$ as

$$\mathcal{B}(y) := \frac{1}{2} \int_{\Omega_{\mathcal{F}}} \left( b_{\mathcal{M}}(y(x)) - b_{\mathcal{F}}(x) \right)^2 \; \mathrm{d}x. \qquad (6)$$

Note that $\mathcal{B}$ coincides with the sum of squared differences of the segmentation masks as binary images. This term has been shown to improve lung boundary alignment, cf. [34].

Although the employed curvature regularization uses second-order derivatives and consequently favors smooth transformations, it cannot safeguard against physically implausible deformations such as large volume expansion or shrinkage and even foldings. This, however, is critical when using properties of the computed transformation such as the Jacobian determinant for the assessment of local lung volume change, cf. [1], [5], [57]–[59].

Consequently, we extend the registration model with an additional term that directly measures change of volume as induced by the transformation $y$,

$$\mathcal{V}(y) := \int_{\Omega_{\mathcal{F}}} \psi(\det \nabla y(x)) \, \mathrm{d}x \qquad (7)$$

with weighting function

$$\psi(t) := \frac{(t-1)^2}{t} \quad \text{for } t > 0 \quad \text{and} \quad \psi(t) := \infty \text{ else,}$$

cf. also [55], [60]. We call $\mathcal{V}$ *volume change control* (VCC). As $\psi(t) = \psi(1/t)$, deviations of the Jacobian from 1, i.e. local volume expansion or shrinkage, are symmetrically penalized. In addition, $\psi$ ensures (local) injectivity of the deformation since $\psi(\det \nabla y) \to \infty$ as $\det \nabla y \to 0$. Hence, $\mathcal{V}(y) = \infty$ if the Jacobian becomes negative at any point. The volume change control therefore fulfills two important functions: It prevents foldings and ensures that large changes in volume are penalized. Our implementation is based on representing transformations as first order B-splines defined on a regular control point grid. In this case, the Jacobian is pointwise bounded by the minimum and maximum of 64 terms that depend on the eight (vector-valued) B-spline coefficients needed to evaluate the B-spline at a single point in 3D [61]. More precisely, we partition $\Omega_{\mathcal{F}}$ into the cells $\Omega_{\mathcal{F}_i}$ of the underlying control point grid with constant volume $C := |\Omega_{\mathcal{F}_i}|$. This enables us to calculate the local Jacobian for all $x \in \Omega_{\mathcal{F}_i}$ as

$$\det \nabla y(x) = \sum_{\ell=1}^{64} \alpha_{i\ell}(x) d_{i\ell} \qquad (8)$$

with $\alpha_{i\ell}(x) \in [0, 1]$, $\sum_{\ell=1}^{64} \alpha_{i\ell}(x) = 1$. The values $d_{i\ell}$ depend on the B-spline coefficients of the transformation $y$ while $\alpha_{i\ell}(x)$ do not. Hence, $\det \nabla y(x)$ is bounded by the minimum and maximum of $d_{i1}, \ldots, d_{i64}$, and with Jensen's inequality we obtain for convex $\psi$

$$\widehat{\mathcal{V}}(y) := C \sum_i \sum_{\ell=1}^{64} \psi(d_{i\ell}) \geq \mathcal{V}(y). \qquad (9)$$

We use $\widehat{\mathcal{V}}(y)$ for our actual implementation. Note that $\mathcal{V}$ is not only bounded by $\widehat{\mathcal{V}}$, but also $\widehat{\mathcal{V}}(y) = \infty$ if $\det \nabla y(x) \leq 0$ at any point $x \in \Omega_{\mathcal{F}}$. Hence, we penalize volume change and numerically guarantee (locally) injective transformations.

*2) Keypoint Integration:* We now describe our approach for adding the sparse keypoint correspondences from Section II-C to the registration model. Generally, the integration of keypoint information into intensity-based image registration can be accomplished in numerous ways, see e.g. [62], [63]. In all these approaches, keypoint matches serve as additional constraints to the sought deformation, either requiring an exact correspondence [62] or allowing for user-defined tolerances [63]. The primary use case of these approaches consists in integrating user input, for which an error estimate is available, into the registration process.

In our algorithm, however, the keypoint input is generated automatically without any user input, and there is no obvious way of deriving reliable upper bounds for the keypoint alignment error. Hence, we choose to integrate the keypoint information through a least-squares penalty into our model, cf. e.g. [24]. This has the additional advantage of a very lightweight computation that can directly be performed within our unconstrained optimization setting.

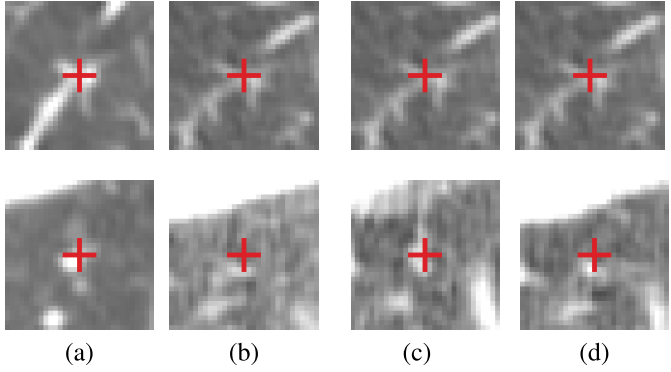

|     |     |     |     |
| --- | --- | --- | --- |
| (a) | (b) | (c) | (d) |

Fig. 3. Axial views of dataset COPD4 [7]: (a) inhale scan with Förstner keypoint location, (b) corresponding location according to a scattered displacement interpolation of nearby manual landmark annotations in exhale scan, (c) correspondence found with keypoint matching, (d) final location after nonlinear registration. The top row shows a normal case where manual annotation, keypoint matching and nonlinear registration are all in agreement. The bottom row demonstrates the ability of DIS-CO to correct an inaccurate keypoint correspondence.

Our aim is to minimize the squared difference between the dense intensity-driven transformation $y$ and the sparse keypoint transformation $y_{KP}$ from Section II-C defined at keypoint locations $x_i$. Naturally, this leads to a discrete sum over all $|K|$ keypoint locations,

$$\mathcal{K}(y) := \sum_{i=1}^{|K|} \|y(x_i) - y_{KP}(x_i)\|_2^2, \quad (10)$$

as an additional element of the joint objective. Fig. 3 qualitatively shows the effect of $\mathcal{K}$. In the top row, the visually correct correspondence from the keypoint matching algorithm is left unaltered by the full method. In the bottom row, an erroneous match (c) is recovered by the deformable registration step (d) as can be seen in comparison to displacement estimates obtained by a scattered interpolation of nearby expert landmarks (b).

With weighting parameters $\alpha, \beta, \gamma, \delta > 0$, the full model is given by

$$\mathcal{J}(y) = \mathcal{D}(\mathcal{F}, \mathcal{M}(y)) + \alpha \mathcal{R}(y) + \beta \mathcal{B}(y) + \gamma \mathcal{V}(y) + \delta \mathcal{K}(y). \quad (11)$$

The parameters $\beta$ and $\delta$ are determined adaptively to balance the influence of the three data terms $\mathcal{D}$, $\mathcal{B}$ and $\mathcal{K}$. Since the intensity-based distance term and the keypoint information are considered to be of equal importance for the alignment, the parameter $\delta$ is chosen such that $\mathcal{D}(\mathcal{F}, \mathcal{M}(y)) = \delta \mathcal{K}(y)$ at the beginning of the registration. Similarly, the influence of the lung boundary alignment term is steered by setting $\mathcal{D}(\mathcal{F}, \mathcal{M}(y)) = 100\beta\mathcal{B}(y)$. As the pre-registration is already designed for precise alignment of the lung boundaries, $\beta$ is decreased by two orders of magnitude in comparison to $\mathcal{D}$; another reason for decreasing $\beta$ is that $\mathcal{D}$ also contributes to the lung boundary alignment.

*3) Numerical Optimization:* The numerical optimization of the joint objective (11) is performed within the discretize-optimize framework [32]. All components of the registration are discretized first, yielding a finite-dimensional optimization problem that can be solved using Newton-type methods. Following [9], we employ the L-BFGS algorithm [64] for minimization. The Hessian approximation is initialized as $R + \tau I$ with $R$ denoting the Hessian matrix of the curvature regularizer and $I$ the identity matrix [32]. For all registrations, $\tau = 10$ and an L-BFGS buffer size of 5 were used.

The numerical optimization is embedded in a multilevel scheme ranging from coarse to fine. Image downsampling is performed after Gaussian smoothing and aiming at isotropic voxels as proposed in [9]. We use a four-level pyramid with the original images on the finest level. A control point grid of size $128 \times 128 \times 128$ is employed on the finest level, the number of control points is halved in each dimension for each coarser level. In our experiments, increasing the number of control points did not further improve accuracy, cf. also [29], [65]. Trilinear interpolation is used to evaluate the deformation at arbitrary positions.

The minimization is performed using the fully matrix-free computation rules for objective function gradient and Hessian approximation as proposed in [56]. These formulations considerably reduce memory footprint, enable better cache efficiency and allow for parallel computations on multicore architectures, resulting in substantially reduced overall execution time. In comparison to [9], we can use denser control point grids with lower memory consumption. For more details, the reader is referred to the description in [66].

## III. EXPERIMENTS AND RESULTS

Several experiments have been performed to assess the accuracy, plausibility and parameter sensitivity of our proposed registration method DIS-CO. The conducted experiments include evaluation of landmark distances on the inspiration-expiration DIR-Lab COPD [7] and 4DCT datasets [67], [68], assessment of the alignment of the major fissures on COPD scans, and a submission to the EMPIRE10 challenge on pulmonary image registration [6]. In addition, the values for the Jacobian determinant of the computed deformation fields are studied, and the sensitivity of the method to variations of its main parameters is analyzed. For all experiments (unless stated otherwise), the main parameters were fixed to $\alpha^{opt} = 2$, $\alpha_{KP}^{opt} = \frac{1}{45}$, $\gamma^{opt} = 0.001$ and $\eta^{opt} = 12$. These values were determined empirically starting from the values given in [9]. All accuracy evaluations exclusively use publicly available data. Fig. 4 shows representative qualitative results for inspiration-expiration registration of two scan pairs from the EMPIRE10 study and of dataset COPD1.

### A. Landmark Distances

It is common practice to measure registration accuracy using expert-annotated landmarks, i.e. at point pairs that have been declared as corresponding to each other by (medical) experts. Our experiments were primarily performed on the DIR-Lab COPD dataset [7], a collection of ten inspiration-expiration cases of the COPDgene study with 300 expert-annotated landmark pairs each. The public availability of the data allows for easy and transparent comparison to published work. The scans have an axial resolution ranging from 0.586 mm × 0.586 mm to $0.742\,\text{mm} \times 0.742\,\text{mm}$ with $512^2$ voxels per slice and a slice thickness of $2.5\,\text{mm}$ with about 120 slices [7].

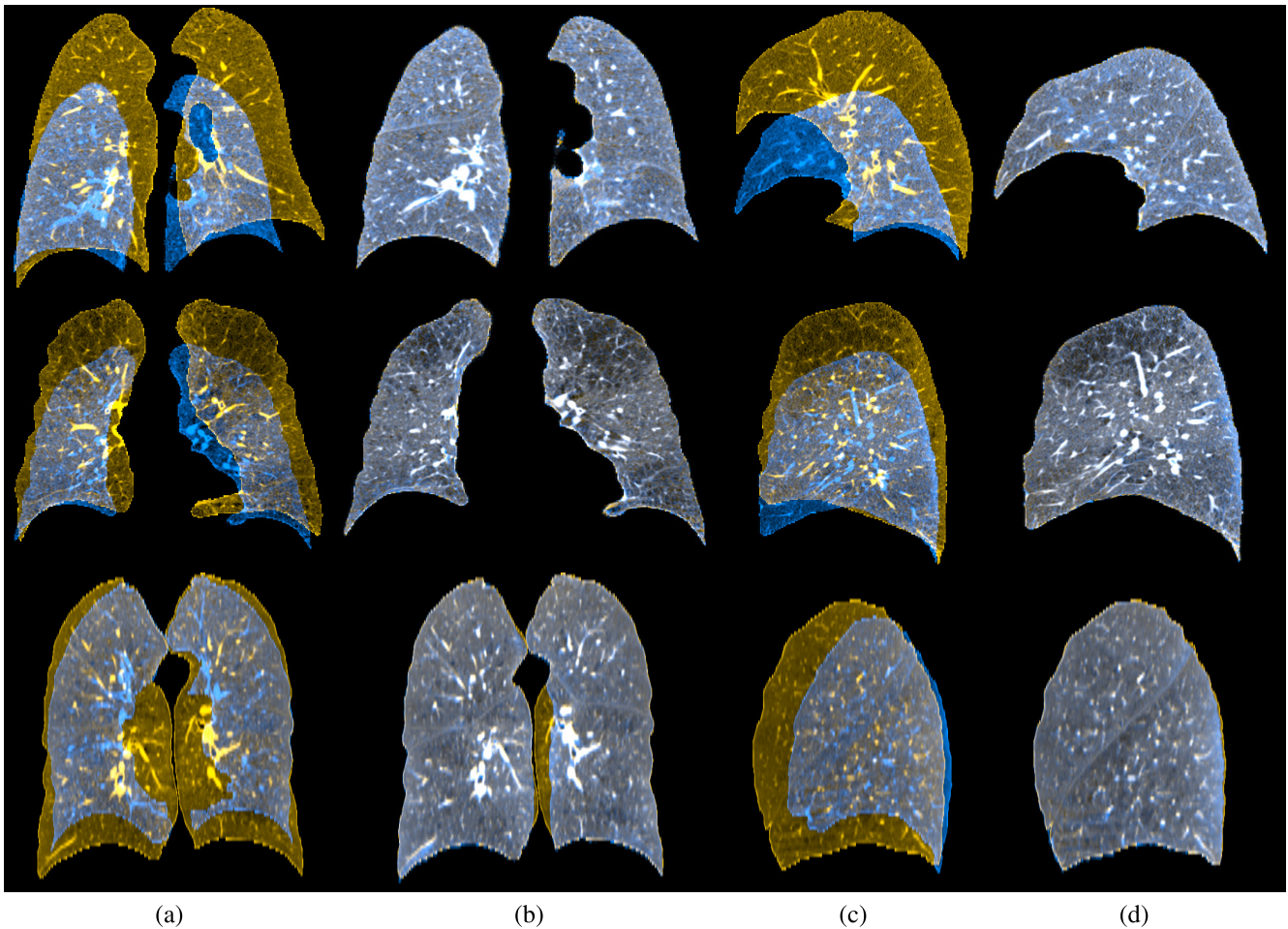

Fig. 4.  Visualization of three inspiration-expiration registration results: (a) and (b) show coronal views before and after registration, (c) and (d) sagittal views, respectively. The color overlays show the inhale scan in orange and the exhale in blue; due to addition of RGB values, aligned structures appear gray or white. The first two rows show the outcome of DIS-CO for the first and seventh scan of the EMPIRE10 challenge, respectively. In both cases the respiratory motion was successfully recovered. The bottom row visualizes the result for the first case of the COPD dataset that reveals a misalignment close to the mediastinum (central region between lungs). This is most likely caused by a joining of left and right lungs during inhale, which would lead to a very high VCC penalty to separate them during registration.

The mean and standard deviation of the distances of the 300 landmarks per dataset after registration were computed for the proposed algorithm and the best performing state-of-the-art methods [8]–[10], [19], [29], [48], [69] with results given in Table I. Here, NLR [9] was adapted to be comparable to our continuous optimization without keypoint matching by using the full image resolution. CORR is the baseline for regularized correspondence search only as originally described in [8]. With an average landmark distance of $0.82 \pm 0.97\,\text{mm}$ after registration, our method outperforms the second best result [69] $(0.96 \pm 1.31\,\text{mm})$ by 15 %. It is herewith the first method to match the inter-observer variability of $0.82 \pm 1.54\,\text{mm}$ as reported for the DIR-Lab COPD dataset [7]. Fig. 5 visualizes the cumulative landmark distance distribution of the compared algorithms. We performed one-sided paired t-tests that show that the improvement to all competing methods is statistically significant, see Table I for p-values.

Since the proposed algorithm employs a Förstner operator-based feature detection, it will preferably select keypoints

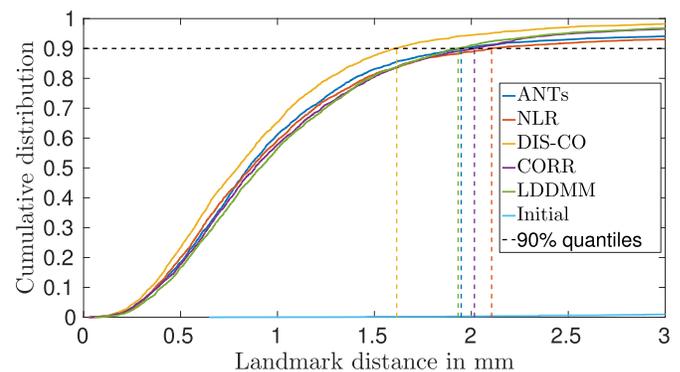

Fig. 5.  Curves of the cumulative distribution of target registration errors for expert landmarks of all COPD datasets after registration. The superiority of our DIS-CO approach is highlighted by the colored dashed lines, which show the 90 % quantile for each method. The quantile values are as follows: 1.61 mm (DIS-CO), 1.93 mm (LDDMM), 1.95 mm (ANTs), 2.02 mm (CORR) and 2.10 mm (NLR).

at positions with high local curvature such as vascular or bronchial bifurcations, cf. [70]. As reported in [7], the manual detection of expert landmarks also typically led to such

| | Initial | SGM [19] | NLR [9] | ANTs [10] | MILO [48] | CORR [8] | LDDMM [29] | isoPTV [69] | DIS-CO | Observer |
|---|---|---|---|---|---|---|---|---|---|---|
| COPD1 | 26.33±11.44 | 1.22±2.73 | 0.92±1.26 | 1.18±1.95 | 0.93±0.92 | 1.00±0.93 | 0.90±0.93 | 0.77±0.75 | 0.79±0.85 | 0.65±0.73 |
| COPD2 | 21.79± 6.47 | 2.48±3.79 | 3.29±5.04 | 2.48±3.82 | 1.77±1.92 | 1.62±1.78 | 1.56±1.67 | 2.22±2.94 | 1.46±2.28 | 1.06±1.51 |
| COPD3 | 12.64± 6.39 | 1.01±0.93 | 0.95±0.94 | 1.00±1.22 | 0.99±0.91 | 1.00±1.06 | 1.03±0.99 | 0.82±0.80 | 0.84±0.82 | 0.58±0.87 |
| COPD4 | 29.58±12.95 | 2.42±3.56 | 0.86±1.00 | 0.98±1.66 | 1.14±1.04 | 1.08±1.05 | 0.94±0.98 | 0.85±0.86 | 0.74±0.86 | 0.71±0.96 |
| COPD5 | 30.08±13.36 | 1.93±3.24 | 1.34±2.41 | 0.93±1.70 | 1.02±1.23 | 0.96±1.13 | 0.85±0.90 | 0.77±0.84 | 0.71±0.83 | 0.65±0.87 |
| COPD6 | 28.46± 9.17 | 1.45±2.42 | 1.94±3.62 | 0.98±1.81 | 0.99±1.08 | 1.01±1.25 | 0.94±1.12 | 0.86±1.12 | 0.64±0.80 | 1.06±2.38 |
| COPD7 | 21.60± 7.74 | 1.05±1.43 | 0.90±1.01 | 1.11±1.81 | 1.03±1.08 | 1.05±1.07 | 0.94±1.25 | 0.74±1.06 | 0.79±0.85 | 0.65±0.78 |
| COPD8 | 26.46±13.24 | 1.16±1.79 | 1.47±2.70 | 1.04±1.74 | 1.31±1.76 | 1.08±1.24 | 1.12±1.56 | 0.81±1.84 | 0.77±0.87 | 0.96±3.07 |
| COPD9 | 14.86± 9.82 | 0.81±0.67 | 0.78±1.17 | 1.02±1.88 | 0.86±1.06 | 0.79±0.80 | 0.88±0.98 | 0.83±1.22 | 0.62±0.66 | 1.01±2.54 |
| COPD10 | 21.81±10.51 | 1.28±1.29 | 0.90±0.88 | 1.57±2.59 | 1.23±1.27 | 1.18±1.31 | 1.17±1.28 | 0.92±0.85 | 0.86±0.88 | 0.87±1.65 |
| Avg. Landm. | 23.36±10.11 | 1.48±2.19 | 1.34±2.00 | 1.23±2.02 | 1.13±1.23 | 1.08±1.21 | 1.03±1.16 | 0.96±1.31 | 0.82±0.97 | 0.82±1.54 |
| Avg. Fiss. | 10.96± 9.93 | - | 1.43±2.81 | 1.47±2.86 | - | 1.21±2.48 | 1.22±2.58 | - | 1.10±2.23 | - |
| *p*-value | 4.4 · 10⁻⁷ | 1.3 · 10⁻³ | 1.3 · 10⁻² | 4.1 · 10⁻⁴ | 1.1 · 10⁻⁵ | 1.2 · 10⁻⁶ | 2.6 · 10⁻⁵ | 5.0 · 10⁻² | - | - |

positions, and consequently there is a potential danger that our algorithm (and similar feature-based methods) put too much focus on these locations. To investigate this, we have performed a second experiment on the DIR-Lab COPD dataset. Here, all keypoints within 10 mm distance from any of the expert-annotated landmarks were removed, and the continuous registration was performed with this reduced set of keypoints only. This procedure only slightly increases the average landmark distance to $0.87 \pm 1.07$ mm, indicating that there is very little bias towards expert-annotated points.

In addition, we have evaluated our algorithm on the widely used DIR-Lab 4DCT dataset [67], [68], a similar collection of ten cases with 300 landmarks on the end-inhale and end-exhale phases of four-dimensional CTs. With the same parameters as for the COPD datasets, we achieve an average landmark distance of $0.94 \pm 1.06$ mm, which is equal to the currently lowest published landmark distances [56] and also very close to the inter-observer variance ($0.88 \pm 1.31$ mm) on this dataset.

### B. Fissure Alignment

The evaluation of registration quality in [6] has shown that fissure alignment is very indicative of the overall performance of different registration approaches. Fissures are very thin anatomical structures that separate individual lung lobes, which are generally difficult to register automatically. They, however, play an important role for analyzing breathing defects. Since no ground truth fissure segmentations were available for the DIR-Lab COPD dataset, we manually segmented the major oblique fissures in all scans in every slice in axial direction.[1] To guide the segmentation in areas of low contrast and to improve consistency across slices, we first performed an initial segmentation in some slices in coronal view. Overall, the task took around 60 minutes per scan pair.

[1] Our segmentations are publicly available at http://mpheinrich.de/research.html#COPD and as supplementary files at IEEE Xplore.

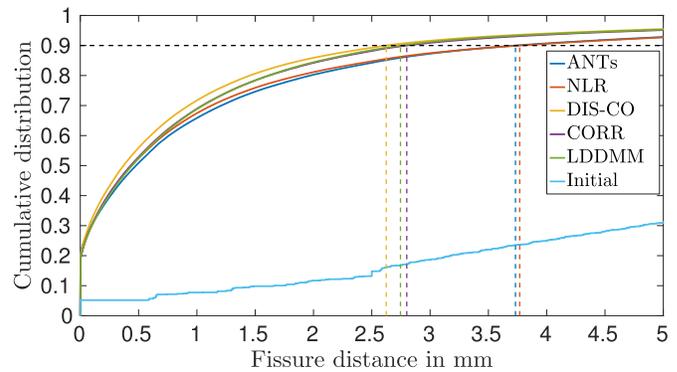

Fig. 6. Curves of the cumulative distribution of distances of fissure voxels in both left and right lung compared to manual segmentations of all COPD datasets after registration. The ranking of different methods is nearly identical to the evaluation based on expert landmarks. The superiority of our DIS-CO approach is highlighted by the colored dashed lines, which show the 90 % quantile for each method. The quantile values are as follows: 2.62 mm (DIS-CO), 2.75 mm (LDDMM), 2.80 mm (CORR), 3.73 mm (ANTs) and 3.77 mm (NLR).

Since the noise level is particularly high in exhale scans, we omitted the horizontal (minor) fissure as also done in [6].

Fissure distances were computed as follows. First, a distance transformation of the labeled moving images was generated. Second, the resulting distance image was transformed using each registration result and linear interpolation. Finally, distances were computed by accessing the voxels of the transformed distance image, which are labeled as fissure in the fixed images. The mean and standard deviation over all datasets is given in Table I. Since this evaluation required access to the full deformation fields, only publicly available algorithms as well as our own previous work could be included into the comparison. The average fissure distances range from 1.10 to 1.47 mm with the proposed method achieving the lowest values. The cumulative distributions of fissure distances after alignment are additionally visualized in Fig. 6.

TABLE II
STATISTICS OF det $(\nabla y)$ WITHIN THE LUNGS FOR THE COPD DATASET.
$Q_1$ AND $Q_{99}$ DENOTE THE 1st AND 99th PERCENTILES. THE NUMBERS
WITHIN PARENTHESES ARE THE RESULTS WITHOUT VCC ($\gamma = 0$).
NO DIFFERENCES IN THE MEAN WERE OBSERVED FOR
THE GIVEN PRECISION.

| # | Mean | Std. Dev. | Minimum | $Q_1$ | $Q_{99}$ | Maximum |
|---|------|-----------|---------|-------|----------|---------|
| 1 | 0.64 | 0.17 (0.23) | 0.11 (-0.56) | 0.27 (0.03) | 1.18 (1.40) | 2.47 (3.50) |
| 2 | 0.79 | 0.32 (0.39) | 0.21 (-0.33) | 0.33 (0.13) | 1.75 (1.92) | 3.39 (4.28) |
| 3 | 0.88 | 0.18 (0.22) | 0.30 (-0.22) | 0.45 (0.34) | 1.30 (1.45) | 2.40 (3.34) |
| 4 | 0.45 | 0.12 (0.18) | 0.08 (-0.51) | 0.30 (0.18) | 0.99 (1.24) | 2.89 (3.18) |
| 5 | 0.55 | 0.17 (0.23) | 0.21 (-0.30) | 0.29 (0.16) | 1.09 (1.30) | 3.36 (3.48) |
| 6 | 0.65 | 0.18 (0.25) | 0.19 (-0.23) | 0.31 (0.14) | 1.20 (1.43) | 3.94 (6.58) |
| 7 | 0.76 | 0.16 (0.20) | 0.27 (-0.22) | 0.47 (0.38) | 1.29 (1.43) | 3.05 (3.57) |
| 8 | 0.52 | 0.15 (0.20) | 0.17 (-0.71) | 0.32 (0.17) | 1.16 (1.38) | 4.29 (4.71) |
| 9 | 0.84 | 0.17 (0.21) | 0.32 (-0.01) | 0.48 (0.41) | 1.24 (1.42) | 2.67 (3.45) |
| 10 | 0.88 | 0.20 (0.30) | 0.29 (-0.27) | 0.38 (0.23) | 1.12 (1.40) | 3.36 (6.33) |

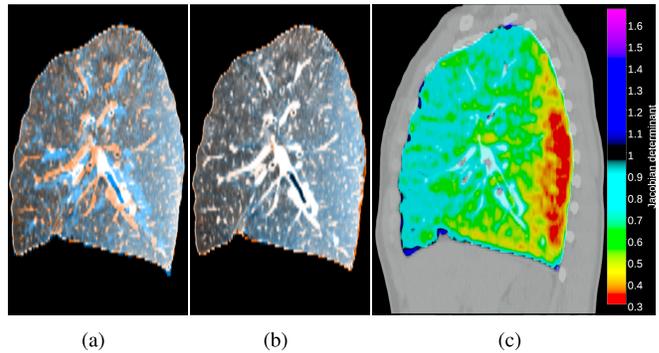

Fig. 7. Sagittal views of dataset COPD6: (a) result of pre-registration, (b) final registration result, (c) resulting Jacobians of the transformation $y$ within the lungs. In (a) and (b) the masked fixed scan is orange and the transformed moving scan is blue; due to addition of RGB values aligned structures appear gray or white.

## C. The EMPIRE10 Challenge

The EMPIRE10 challenge [6] is currently the most comprehensive public comparison study on pulmonary image registration worldwide. Since its start in 2010, 42 lung registration algorithms from academia and industry have been submitted for public evaluation and comparison. The evaluation data consists of 30 datasets from six different categories: repeated breathhold inspiration, breathhold inspiration and expiration, 4D data, ovine data, contrast-noncontrast, and artificially warped scan pairs. Algorithms are evaluated using four different criteria: lung boundary alignment, major fissure alignment, distance of expert-annotated landmarks, and deformation field singularities. To account for the high importance of lung boundary alignment, we increased the weighting of $\mathcal{B}$ tenfold for our submission. All other parameters were left unchanged. As of September 2016, our algorithm ranks first in the EMPIRE10 challenge with an average landmark distance of 0.63 mm. For full details, we refer to the challenge website.[2]

## D. Jacobian Determinant

As motivated in the paragraph on volume regularization, the registration should not produce transformations with a negative Jacobian determinant. Furthermore, very large values or values close to zero are unrealistic as the Jacobian determinant is a local measure for volume change, i.e. if $\det(\nabla y) > 1$ a volume expansion occurred and if $\det(\nabla y) < 1$ the volume decreased. We evaluate the Jacobian determinant on the DIR-Lab COPD dataset using several statistics. The standard deviation of the Jacobian determinant is often used to describe the smoothness of a transformation, see e.g. [20], [57]. The results for the proposed method DIS-CO are given in Table II (first values per column). Note that the evaluation was done in the lung volumes only as we perform masked registrations and the values outside of the lung volume are thus not meaningful (although they never featured negative Jacobian determinants).

On average, a Jacobian less than 1 is expected as $y$ describes the change from full breathhold inspiration to expiration state. For each dataset, this criterion was fulfilled by our algorithm.

In addition, the minimum values are always positive as guaranteed by the volume penalty term. The 1 % and 99 % quantiles as well as the minimum and maximum of the Jacobian show that the vast majority of volume changes is plausible. Fig. 7 shows the inspiration-expiration registration of dataset COPD6 including the pre-registration, the final result and the Jacobian determinant of the computed transformation.

## E. Influence of Additional Objective Function Terms

The objective function of the proposed registration approach is composed of the standard model $\mathcal{D}(\mathcal{F}, \mathcal{M}(y)) + \alpha\mathcal{R}(y)$ and three additional terms $\mathcal{B}$, $\mathcal{V}$ and $\mathcal{K}$. The individual influence of these terms is assessed on the basis of the DIR-Lab COPD datasets. Since we are interested in the contribution of the individual terms, only one term is removed at a time by setting its weight to zero. All other parameters are left unchanged.

*1) Lung Boundary Alignment:* The purpose of the term $\mathcal{B}$ is to improve the alignment of the lung boundaries. Since the nonlinear pre-registration and the term $\mathcal{B}$ are jointly designed for this goal, we additionally changed the pre-registration to a purely affine-linear transformation model for this experiment. The effect of the boundary alignment strategy is analyzed using the Hausdorff distance of the lung segmentations. Without $\mathcal{B}$, an average Hausdorff distance of 11.72 mm is obtained as opposed to 10.20 mm with the full model. In addition, the average landmark distance slightly increases from 0.82 mm for the full model to 0.86 mm. The boundary alignment term $\mathcal{B}$ and the nonlinear pre-registration thus moderately improve both the registration of lung surface and inner-lung structures.

*2) Volume Change Control (VCC):* The aim of using the volume change control mechanism $\mathcal{V}$ is to prevent foldings and implausible volume change. We therefore compare statistics of the Jacobian determinant of transformations with and without VCC, results are given in Table II. Without VCC, foldings are present in all ten cases. In addition, the transformations exhibited a much larger range of volume change and an increased standard deviation of the Jacobian (about 25 % on average) as opposed to the full model. This confirms that VCC produces smoother and more realistic transformations, which is important for the analysis of COPD. Furthermore,

TABLE III

Landmark Error Results of the Parameter Sensitivity Tests on the COPD Datasets [7]. The Resulting Mean Landmark Errors Over All Ten Registrations Are Shown When Varying One Parameter by Orders of Magnitude at a Time and Fixing the Others to Their Empirically Determined Optimal Values. The Colormap Indicates Excellent (Green) to Poor (Red) Quality.

| optimal value → | $\alpha = 2$ | $\alpha_{\mathrm{KP}} = \frac{1}{45}$ | $\gamma = 0.001$ | $\eta = 12$ | colormap |
|---|---|---|---|---|---|
| $\times 10^{-5}$ | 0.88 | 10.71 | 1.31 | 1.67 | 0.80 |
| $\times 10^{-4}$ | 0.92 | 10.71 | 1.19 | 1.22 | 0.95 |
| $\times 10^{-3}$ | 0.88 | 10.75 | 1.01 | 1.1 | 1.10 |
| $\times 10^{-2}$ | 0.87 | 10.58 | 0.95 | 0.91 | 1.25 |
| $\times 10^{-1}$ | 0.85 | 0.94 | 0.86 | 0.86 | 1.40 |
| optimal | 0.82 | 0.82 | 0.82 | 0.82 | 1.55 |
| $\times 10^{1}$ | 1.02 | 0.91 | 1.02 | 1.31 | 1.70 |
| $\times 10^{2}$ | 1.55 | 1.63 | 4.85 | 11.26 | 1.85 |
| $\times 10^{3}$ | 2.82 | 4.94 | 8.03 | 13.17 | 2.00 |
| $\times 10^{4}$ | 4.6 | 5.68 | 7.97 | 13.37 | 2.15 |
| $\times 10^{5}$ | 6.69 | 5.64 | 7.46 | 13.33 | ≥2.30 |

the landmark distance substantially increased to 1.45 mm when omitting the volume change control term.

*3) Keypoint Matching:* The keypoint matching term $\mathcal{K}$ is designed to increase registration accuracy in the case of large motion and to avoid local minima during optimization. When excluding this term, we observed that for five out of the ten DIR-Lab COPD cases (1, 2, 5, 6 and 8), the average landmark distance after registration substantially increases by 2.04 mm on average. For the remaining five cases, only a small decrease in registration accuracy by less than 0.1 mm was found. In total, the overall landmark distance of 0.82 mm for the full model more than doubles to 1.86 mm without $\mathcal{K}$. This clearly demonstrates that the keypoint information helps to increase registration accuracy as validated by 3000 expert landmark pairs. A similar observation was made for the fissure matching, where the average fissure distance over all cases increases by almost 1 mm to 2.05 mm when turning off the keypoint term.

### F. Parameter Sensitivity

We analyzed the sensitivity of the proposed method to variations of its major parameters. As the weights $\beta$ and $\delta$ are chosen adaptively, the proposed method has four main parameters: $\alpha$, $\alpha_{\mathrm{KP}}$, $\gamma$ and $\eta$. Their influence on the registration accuracy is studied by varying one parameter at a time, multiplying it with $\{10^k | k = -5, -4, \ldots, 5\}$, and fixing the remaining three to their empirically determined optimal values. Results are shown in Table III, demonstrating good robustness. $\alpha$, $\gamma$ and $\eta$ can be decreased by up to five orders of magnitude with only moderate impact on registration quality. When the regularization parameters are increased by more than a factor of ten, the registration is substantially deteriorated, which can be explained by the increased stiffness of the transformation model. A similar behavior can be observed for unreasonably large edge parameters for NGF. When $\alpha_{\mathrm{KP}}$ is multiplied with values equal to or less than $10^{-2}$ the accuracy is substantially reduced, underlining the importance of the regularization in the keypoint detection. In addition, a two-fold cross-validation using a moderate-sized grid search was explored as alternative to our heuristic parameter optimization. This yields slightly different optima for $\alpha_{\mathrm{KP}}$ and an average error of 0.83 mm.

## IV. DISCUSSION AND CONCLUSION

We have presented a new deformable registration approach targeted to the challenging alignment of inspiration and expiration scans of lung patients. While most state-of-the-art algorithms struggle with the large magnitude of this highly deformable motion, we demonstrate an excellent alignment of lung vessels, airways and fissures, while constraining the transformation to be physically plausible (with positive Jacobians) and accurately match the lung boundaries. Our target registration error of 0.82 mm for expert-annotated landmarks on the DIR-Lab COPD database sets a new state-of-the-art accuracy (outperforming all previous approaches by 15 %) and reaches the inter-observer variability. Our method also ranks first overall in the comprehensive EMPIRE10 lung registration challenge [6]. In addition, the computation time of less than five minutes on a standard multi-core computer would enable practical implementation in clinical routine.

This registration performance is achieved by combining three key elements. First, the robustness against large motion between breathing states is substantially improved by finding sparse regularized keypoint correspondences using a dense sampling of potential displacements together with a part-based model regularization. Second, a volume change control mechanism is employed that reliably guarantees a positive Jacobian (no folding) and enables a low-complexity transformation, while a lung mask term ensures alignment of the outer boundaries. Third, the use of quasi-Newton methods in a matrix-free setting enables us to substantially improve computational efficiency. All parts of our method are integrally combined within one unified registration framework that optimizes the same similarity cost function in both discrete and continuous matching. Further improvements could be achieved by learning the most discriminative descriptor or similarity metric for each stage, cf. [44]. However, substantially different cost functions could also lead to contradictory objectives, which would at least delay convergence, and thus have to be carefully studied prior to combined usage.

We performed an extensive evaluation on ten inhale-exhale CT scan pairs of the COPDgene study, ten 4DCT scans of the DIR-Lab database and all 30 datasets of the EMPIRE10 challenge, which cover a variety of further application scenarios. We provide a detailed sensitivity analysis that demonstrates the robustness of the method against parameter variations. Furthermore, we have manually segmented 40 lung fissures (both oblique fissures in all COPD scan pairs) and provide them as download for interested researchers as further validation metric. A low surface distance of fissures after alignment, as achieved by our approach, is important when using deformable image registration for assessing breathing defects (reduced lung ventilation and therefore low volume change [3]) for individual lobes. While the treatment of sliding motion was not a focus of this work, the discrete keypoint matching in principle allows for accurate registration both within and outside the lungs. However, a more adaptive, non-quadratic regularization model, e.g. a total variation term as proposed in [69], would have to be adopted in the deformable registration step. In particular, the combination with the volume change control mechanism is not straightforward.

Our quantitative results clearly demonstrate the usefulness of integrating regularized correspondences into a continuous registration framework. Our method combines high alignment accuracy with physiologically plausible transformations at a moderate, clinically acceptable runtime. We hope that our contribution will further advance the application of pulmonary image registration for local lung ventilation assessment and radiation therapy planning in the clinical practice.

## ACKNOWLEDGMENT

The authors are deeply grateful to Keelin Murphy, Edward Castillo and Richard Castillo for providing evaluation benchmarks for pulmonary image registration.